# TRANSFERABILITY ANALYSIS OF DATA-DRIVEN ADDITIVE MANUFACTURING KNOWLEDGE: A CASE STUDY BETWEEN POWDER BED FUSION AND DIRECTED ENERGY DEPOSITION

**Mutahar Safdar**
McGill University
Montreal, QC,
Canada

**Jiarui Xie**
McGill University
Montreal, QC,
Canada

**Hyunwoong Ko**
Arizona State University,
AZ, USA

**Yan Lu**
National Institute of
Standards and
Technology
MD, USA

**Guy Lamouche**
National Research Council
Montreal, QC, Canada

**Yaoyao Fiona Zhao***
McGill University
Montreal, QC, Canada

## ABSTRACT

*Data-driven research in Additive Manufacturing (AM) has gained significant success in recent years. This has led to a plethora of scientific literature to emerge. The knowledge in these works consists of AM and Artificial Intelligence (AI) contexts that haven't been mined and formalized in an integrated way. Moreover, no tools or guidelines exist to support data-driven knowledge transfer from one context to another. As a result, data-driven solutions using specific AI techniques are being developed and validated only for specific AM process technologies. There is a potential to exploit the inherent similarities across various AM technologies and adapt the existing solutions from one process or problem to another using AI, such as Transfer Learning. We propose a three-step knowledge transferability analysis framework in AM to support data-driven AM knowledge transfer. As a prerequisite to transferability analysis, AM knowledge is featurized into identified knowledge components. The framework consists of pre-transfer, transfer, and post-transfer steps to accomplish knowledge transfer. A case study is conducted between flagship metal AM processes. Laser Powder Bed Fusion (LPBF) is the source of knowledge motivated by its relative matureness in applying AI over Directed Energy Deposition (DED), which drives the need for knowledge transfer as the less explored target process. We show successful transfer at different levels of the data-driven solution, including data representation, model architecture, and model parameters. The pipeline of AM knowledge transfer can be automated in the future to allow efficient cross-context or cross-process knowledge exchange.*

## 1. INTRODUCTION

Additive Manufacturing (AM) or three-dimensional (3D) printing is used to fabricate parts layer-wise as opposed to the subtractive approach of material removal. The American Society of Testing Materials (ASTM) defines seven standardized categories of AM processes [1]. These technologies can support various applications such as tool elimination, material savings, design freedom, cost reduction, part consolidation, prototyping ease, mass customization, and production efficiency. The advantages of AM have led to increased attention from academia and industry to advance AM technologies for even broader applications and ultimately mature them to rival conventional manufacturing methods at the industrial scale.

Metal AM (MAM) can manufacture fully dense metallic parts directly from 3D digital designs. Though these processes provide much more design freedom and lead to material saving, they encounter distinct challenges of process control and part reproducibility due to their unique nature of lengthy material joining and part fabrication. Laser Powder Bed Fusion (LPBF) and Directed Energy Deposition (DED) are two representative MAM technologies, each providing unique benefits. LPBF uses

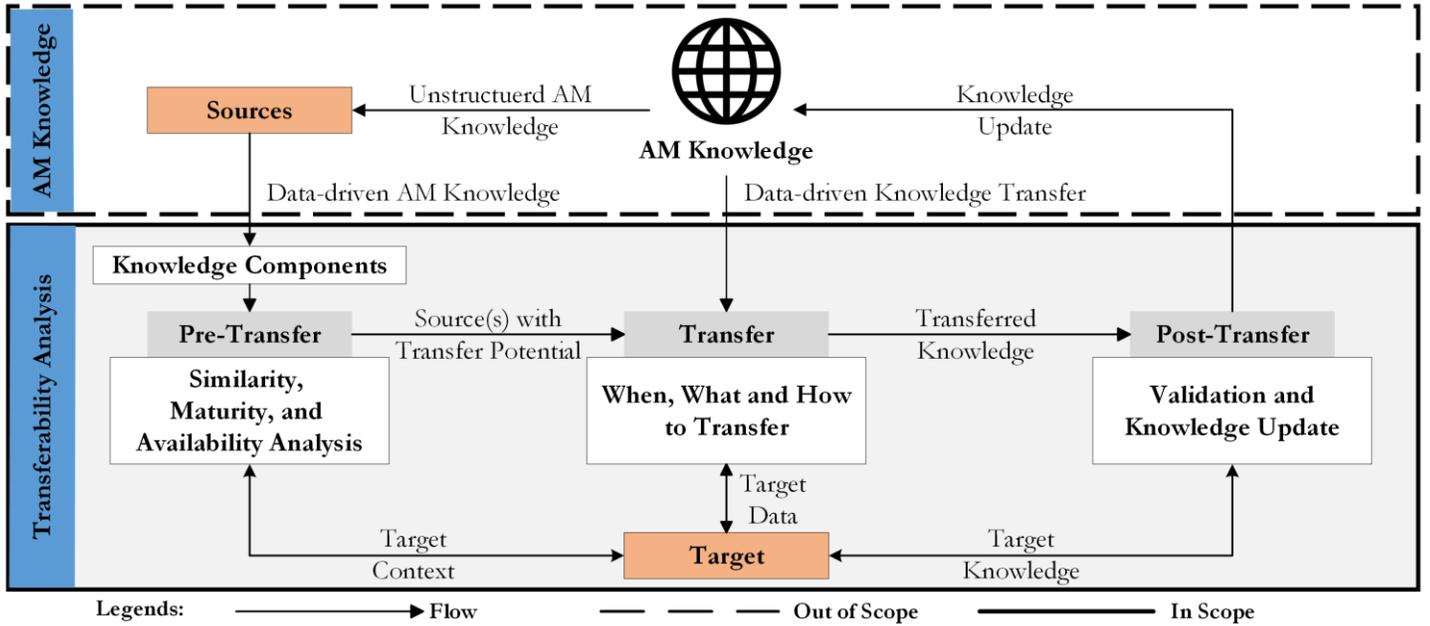

**FIGURE 1:** THREE-STEP KNOWLEDGE TRANSFERABILITY ANALYSIS FRAMEWORK. THE ORANGE SHADE REPRESENTS EXISTING SOURCE AND TARGET KNOWELDGE WHERAS THE GREY SHADE HIGHLIGHTS STEPS OF THE FRAMEWORK

an energy source to fuse pre-laid powder layers leading to a 3D part upon successive repetition of the process. DED also uses a focused energy source as the material is supplied on the fly. Both processes have advantages and disadvantages, suiting diverse applications, but share similar part quality control challenges.

To enhance process control and part reproducibility in MAM and other AM processes, various analytical, numerical, and empirical solutions are being developed independently, and usually toward a specific process, a combination of materials, layering, and machine technology. The resulting knowledge is only validated for the specific process. For instance, the recent wave of Machine Learning (ML) aided research in AM has introduced data-driven solutions to solve different problems of each process [2]. Some of these frameworks and solutions are expected to expedite AM development significantly. For example, co-authors of this paper developed ML models on melt-pool data to enhance in-situ monitoring and control of LPBF [3, 4]. In [3, 5], the co-authors also presented ML-based methods to extract new process-structure-property causal knowledge from AM data to enhance control activities and part reproducibility. [2, 6, 7] present more ML approaches in AM in their literature reviews.

There is a significant research problem identified from the literature reviews - the ML solutions of many existing AM studies are only developed and verified for specific AM processes. While research on different aspects of AM knowledge and its management exists [3, 5, 8, 9], no framework supports cross-process knowledge sharing. To address the challenge, this paper proposes a novel Transfer-Learning (TL)-based framework for knowledge transferability analysis in AM. This paper also presents a case study demonstrating the framework using LPBF and DED data.

The remainder of the paper is as follows. Section 2 introduces the framework. Section 3 demonstrates the case study. Section 4 presents the results and discussion. We conclude this article with concluding remarks and future work in Section 5.

## 2. FRAMEWORK FOR TRANSFERABILITY ANALYSIS

We present a three-step framework to support the transferability analysis of AM knowledge between two flagship process categories: LPBF and DED. Figure 1 introduces the framework indicating the steps pre-transfer, transfer, and post-transfer involved in the process of knowledge transfer. The knowledge components are identified as a pre-requisite to transferability analysis. The proposed framework is independent of the nature of a source solution and provides a generic approach to conduct knowledge transfer across ML solutions.

### 2.1 Knowledge Components

While the representation and management of ML-aided AM knowledge is considered out of the scope, a criterion to represent AM knowledge is needed within the scope of transferability analysis. We featurize ML-aided AM knowledge into AM and ML knowledge components in support of transferability analysis. This step serves as the pre-requisite to the proposed framework. The components are chosen to cover varying levels of domain knowledge (both AM and ML) and are arranged according to their significance in each domain.

The ordinal levels of AM knowledge are defined below:
*AM Process (AM_P)*: The AM process types are identified as the first level of AM knowledge. This is where different categories of AM significantly vary from each other.
*AM Material (AM_MT)*: The AM material types fall at the second level of AM knowledge. Since different material systems represent different knowledge developed for the same process, it is important to distinguish each AM activity in terms of the specific material type used. Some materials may be more developed than others (e.g., Ti4Al6V), leading to high chances of knowledge transfer.

*AM System (AM_S)*: The AM system setups for a specific AM process type are identified as the third level of AM knowledge. A system refers to both base printer setup and added hardware (e.g., sensors) supporting the ML solutions.

*AM Model (AM_M)*: An AM model refers to the developed knowledge specific to a given process, material, and system. Over the past decade, researchers have developed and proposed numerous AM models (e.g., analytical, numerical).

*AM Activity (AM_A)*: AM activity is the focus of each data-driven solution and lies at the fifth level of AM knowledge. A full or partial overlap in the preceding levels can open possibilities of activity-based knowledge transfer. This level organizes characteristics according to the lifecycle of AM, namely design, process, and product-based activities.

*AM Concern (AM_C)*: AM concern is identified as the sixth and final level of AM knowledge. Design concerns can have types such as manufacturability or deviation prediction. Processes can be divided into different states, such as normal and abnormal. Products can be judged against various quality metrics, such as macro and micro mechanical defects. A match at the activity level doesn't imply a full match until a specific type of AM concern is found to be the focus of said AM activity.

The key levels of ML knowledge are also identified ordinally:

*ML Task (ML_T)*: An ML task is identified as the first level of ML knowledge. ML tasks are divided into broad categories of regression, classification, and clustering. ML applications in AM can be first arranged into specific tasks indicating maximum variation of contained knowledge.

*ML Model (ML_M)*: An ML model represents the second level of ML knowledge. For a given task, a multitude of ML models exists to either learn input-output relations or discover underlying patterns. It is important to identify the model type used in each application for knowledge transfer. Shallow and deep models are representative of two major categories of ML models. However, specific types of models (e.g., Linear Regression, Support Vector Machines, Convolutional Neural Network or CNN) are used in practice.

*ML Input (ML_I)*: ML input stands at the third level of ML knowledge in AM applications. Empirical models can unearth correlations for a characteristic of interest using different types of inputs. Based on trends in AM, input types are divided into Graphic, 3D, Tabular, and Sequence [10].

*ML Preprocessing (ML_P)*: ML preprocessing is added as the fourth level of ML knowledge. Data handling techniques refer to any action that leads to improvement in data quality for a given learning task.

*ML Output (ML_O)*: ML output lies at the end of ML knowledge context. Identifying the nature and availability of outputs is important in knowledge transferability scenarios and helps to compare a label space during the task comparison of the transfer steps.

## 2.2 Pre-Transfer

Once an instance (or several instances) of a target context is selected, the pre-transfer, which begins the transferability analysis steps, is followed to conduct knowledge transfer. The main objective of the pre-transfer step is to comprehensively compare source and target knowledge components, identify the similarity between the components, and select the potential sources that can be used to transfer the knowledge to a target's context. This step is also aimed at the applicability analysis of a given source knowledge to a target context.

The similarity is defined as the one-to-one similarity of AM and ML knowledge in terms of their components. Each one-to-one comparison leads to a similarity/applicability (1) or dissimilarity/inapplicability (0) score. All scores are added and normalized to get a final similarity index in the range of 0 to 1.

The pretransfer step also performs a maturity check on source knowledge solely. The maturity is defined in terms of the newness and performance of the knowledge being transferred. The metric for performance depends on the ML task, while the newness is indicated by the robustness of the approach and its history of successive development. The performance metric gets a value of 1 for the highest reported performance and is successively reduced by 0.1 for sources with low performance. The newness metric is assigned 1 for the latest reported work and gets lowered by 0.1 for previously reported sources in the order of publication. The maturity factor is defined as the sum of performance and newness metrics and is later normalized in the range of 0 to 1.

**TABLE 1:** PRE-TRANSFER ANALYSIS

| Similarity Analysis | Maturity Analysis | Availability Analysis |
|---|---|---|
| AM Similarity Score = $\sum S_{AM}$<br>ML Similarity Score = $\sum S_{ML}$<br>Knowledge Components = $\sum KC_t$ | Newness Score = $n$<br>Performance Score = $p$ | Available Knowledge = 1<br>Unavailable Knowledge = 0 |
| Similarity Index (S) = $(\sum S_{AM} + \sum S_{ML})/\sum KC_t$ | Maturity Factor (M) = $(n+p)/2$ | Availability Factor (A) = 0 or 1 |
| Pre-Transfer Score = $S \times M \times A$ | | |

Before knowledge can be transferred, all components of the knowledge from AM and ML are needed. Availability check results in 1 for available and 0 for unavailable knowledge. Both maturity and availability results are factored in the similarity index to get the results from the pre-transfer analysis. The availability of knowledge (especially data and models) could be a major bottleneck to knowledge sharing in AM as datasets, and specific solutions are often not openly available for re-use and transfer. Table 1 highlights sub-analysis steps for pre-transfer leading to a pre-transfer score indicative of knowledge transfer potential.

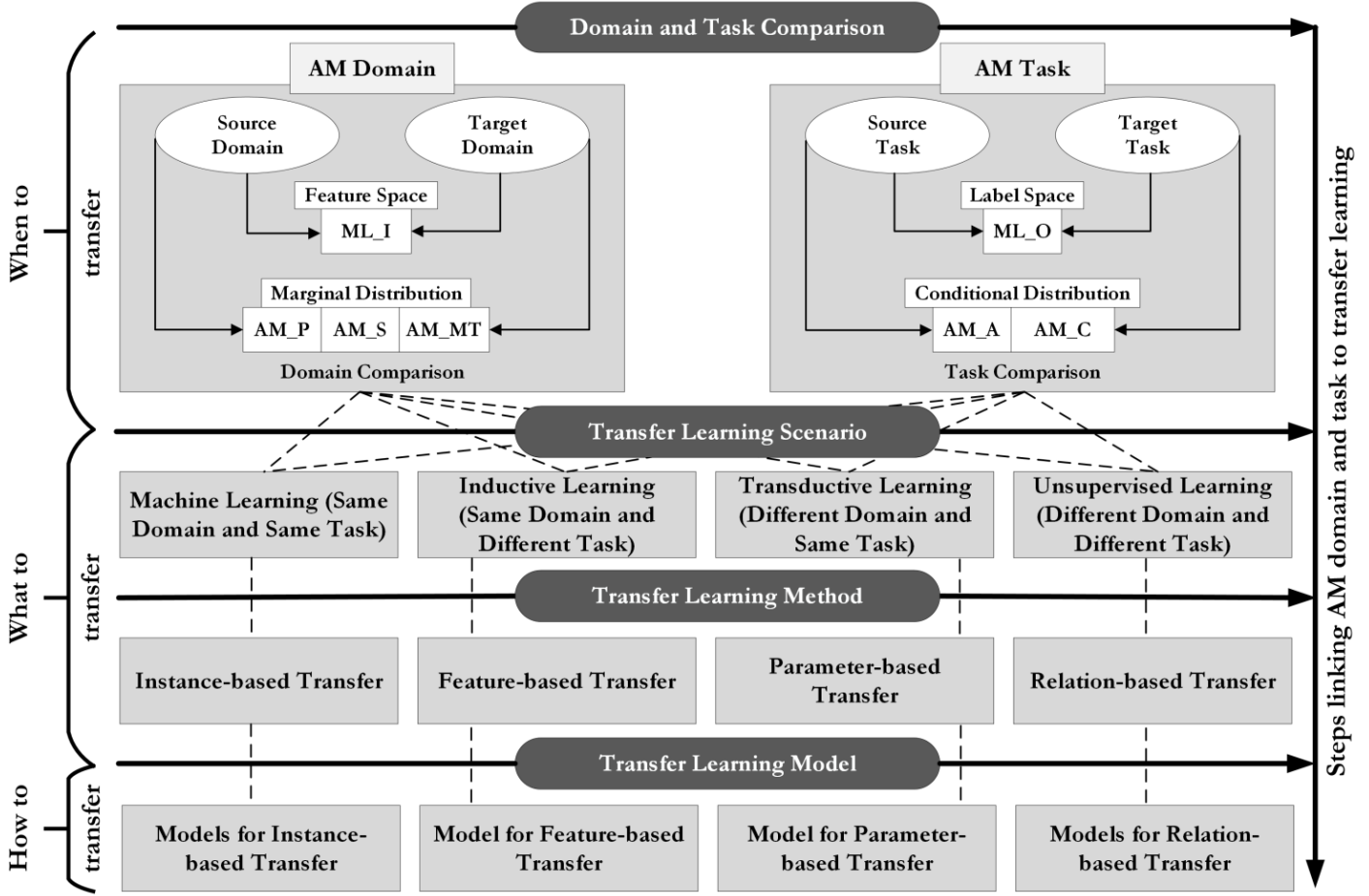

**FIGURE 2:** DOMAIN AND TASK COMPARISON TO INDENTIFY TRANSFER LEARNING SCENARIO, METHOD, AND MODELS. ML_I AND ML_O REFER TO ML INPUT AND ML OUTPUT WHERAS AM_P, AM_S, AM_MT, AM_A, AND AM_C REFER TO AM PROCESS, AM SYSTEM, AM MATERIAL, AM ACTIVITY, AND AM CONCERN RESPECTIVELY.

The highest possible score of 1 from the pre-transfer step represents an ideal match with the same process. For cross-process knowledge transfer, the source context with the maximum score can be considered for the transfer step in the case of several potential sources. A score greater than 0.5 represents a significant overlap of a target context with mature and available source knowledge.

**2.3 Transfer**
The pre-transfer analysis leads to the identification of potential knowledge sources and helps to quickly filter and arrange them based on the assigned pre-transfer scores. However, there is no guarantee that these can support knowledge transfer, as some scenarios can be without any meaningful overlap (e.g., scenarios with zero ML similarity scores). Nonetheless, some knowledge may still be transferable such as specific experimental design methods or AM hardware used to generate data. As such, the non-ML solutions to knowledge transfer can be straightforward. We focus on ML-based solutions for AM knowledge transfer. The ML-based solutions can be grouped under the umbrella term of TL and are identified systematically in Figure 2.

The pre-transfer analysis can lead to different types of similarity scenarios, such as full AM similarity, full ML similarity, partial AM similarity, partial ML similarity, and mixed AM and ML similarity. To link these with a potential TL model, we propose to translate the identified similarity scenario to a TL scenario using TL knowledge. TL is motivated by the fact that learning done for a previous and similar task can expedite the learning for a related new task in the absence of sufficient high-quality data. In their representative survey on TL, Pan and Yang gave a generic definition of transfer learning as [11]: "Given a source domain $D_s$ and learning task $T_s$, a target domain $D_t$ and learning task $T_t$, transfer learning aims to help improve the learning of the target predictive function (Ft(.)) in $D_t$ using the knowledge in $D_s$ and $T_s$, where:"

$$D_s \neq D_t \tag{1}$$

or,

$$T_s \neq T_t \tag{2}$$

The AM similarity scenarios are represented in terms of domains and tasks from source and target contexts. They define a domain in terms of a feature space $X$ and a marginal probability distribution $P(X)$ as

$$\mathcal{D} = \{X, P(X)\} \quad (3)$$

where X= $\{x_1, x_2, x_3, ..., x_n\} \in X$

For two domains to be considered similar, their feature spaces and marginal distribution should match. A feature space is representative of all the features in each domain, while a marginal probability distribution is the probability of seeing a specific instance of the feature in that domain. Examples of a feature space can be a data representation such as an image format or the language of a text. Whereas a marginal distribution will be the probability of specific features (e.g., pixels, words) in each space.

For a given domain, a task is defined by a label space $Y$ and a conditional probability distribution $P(Y|X)$ as

$$\mathcal{T} = \{Y, P(Y|X)\} \quad (4)$$

where Y= $\{y_1, y_2, x_3, ..., y_n\} \in Y$

For two tasks to be considered similar, both label space and conditional distributions should match. A label space is representative of all labels for a given task, while conditional probability is the probability of seeing a specific label against a specific instance of the input. Examples of a label space include anomaly prediction, binary, or multi-class label spaces. A conditional probability will be one specific label value for a given instance in the feature space.

The domain and task for source and target AM context are defined in terms of the knowledge components identified earlier and as shown in Figure 2. An AM domain for TL can be defined based on AM_P, AM_S, AM_MT, and ML_I types where ML_I represents feature space and remaining parameters determine the marginal distribution of features in the space as

$$\mathcal{D}_{AM} = \{X_{ML\_I}, P(X_{ML\_I})\} \quad (5)$$

Similarly, an AM task for TL can be defined based on AM_A, AM_C and associated ML_O where ML_O defines the label space and the remaining parameters determine the conditional distribution of labels as

$$\mathcal{T}_{AM} = \{Y_{ML\_T}, P(Y_{ML\_T}|X_{ML\_I})\} \quad (6)$$

Once the AM source and target contexts are discretized into domain and task, the overall process of knowledge transfer revolves around answering three key questions as below:

*When to transfer*: This is the first step in Figure 2. Domain and task comparison between source and target scenarios is the main step to answering, "when to transfer?" question. The comparison between source and target can lead to four broad outcomes as: Same domain with same task, same domain with different task, different domain with same task, and different domain with different task. This sets the stage for knowledge transfer and helps answer the remaining questions. The first outcome represents a traditional machine learning problems where nothing is different, and source represents training setting while target represents test setting. The second outcome is referred to as inductive transfer learning. Based on the nature of task difference (label space or conditional probability distribution), an appropriate TL method and model can be selected. The third outcome where task is same for source and target is referred to as transductive transfer learning. Finally, the situation where both domain and task differ is referred to as unsupervised transfer learning.

*What to transfer*: Identification of a TL scenario helps select an appropriate TL method and answers "what to transfer" question, identified as the second step in Figure 2. Bang et al. surveyed TL methods and evaluated their applicability in manufacturing scenarios [12]. Their approach to select TL method combines TL scenario with the availability of labels in the target domain. A similar approach is worth considering in AM since it's difficult to obtain sufficient labeled data in industrial settings. A recent text on the topic classified TL methods into four main categories namely instance-based transfer, feature-based transfer, parameter-based transfer, and relation-based transfer [13].

*How to transfer*: Different models are available for each TL method. TL model development steps involve iterations over different model types to maximize transfer of mined knowledge. Figure 2 indicates that TL models can be seen to fall into the categories of instance-based, feature-based, parameters-based, and relations-based.

**2.4 Post-Transfer**

The post-transfer step is simple as compared to the previous two steps. It involves validation of the transferred knowledge. In addition to testing on the target data post knowledge transfer, this step also involves the update of existing AM knowledge depending on the results of the transfer process. The post-transfer step can also involve rigorous testing on the limits of knowledge transfer once the initial TL method yield sufficient results. This can help determine the limits on the target knowledge (e.g., dataset size) that can yield acceptable performance with the available source data and model.

**3. CASE STUDY**

The case study is conducted between LPBF and DED processes that have been published in the past. Specifically, the LPBF process and associated model come from the National Institute of Standards and Technology (NIST) and dataset is openly available [14]. The DED dataset is from Mississippi State University (MSU) without any associated model or knowledge [15]. The rationale is that source model and knowledge can be re-used following the presented framework for the task at hand. The task considered for the target context is anomaly detection based on process data.

**TABLE 2:** PRE-TRANSFER ANALYSIS BETWEEN SOURCE AND TARGET KNOWLEDGE COMPONENTS

| Knowledge Component | Source Context | Target Context | Pre-Transfer Observation (Score) |
|---|---|---|---|
| *AM Knowledge Components* | | | |
| *AM Process* | LPBF: The LPBF process is used to fuse pre-laid layers of powder using an energy source | DED: The DED process uses an energy source to fuse materials as they are being deposited. | Both AM processes differ leading to change in marginal distribution (0) |
| *AM Material* | Substrate: Wrought nickel alloy 625 (IN625) Feedstock: Recycled IN625 powder | Substrate: Ti4Al6V Feedstock: Ti4Al6V powder | Both AM materials differ leading to change in marginal distribution (0) |
| *AM System* | Base System: Additive Manufacturing Metrology Testbed (AMMT) at NIST Camera Specifications: - Visible Light Camera - CMOS Detector - 120 by 120 pixels window - 8 um pixel pitch - Frame Rate 10000 Hz | Base System: OPTOMEC LENS™ 750 system equipped with a 1 kW Nd:YAG laser (IPG) at MSU Camera Specifications: - Dual Wavelength Pyrometer - CMOS Detector - 752 by 480 pixels window - 6.45 um pixel pitch - Frame Rate 6.4 Hz | The systems used were different. The data collected represents melt pool of both processes, each with different technical camera specifications. The process monitoring data type was same (e.g., graphic or pixel topology data)., representing a similarity of feature space for ML input. (0) |
| *AM Model* | A spatiotemporal model for process data representation | None | The source AM model can be adapted to target's context to represent and arrange data (1) |
| *AM Activity* | Process | Process | Both contexts focus on same level of AM characteristic (1) |
| *AM Concern* | Process Anomaly Detection | Process Anomaly Detection | The type of process characteristic in source and target is same. (1) |
| *ML Knowledge Components* | | | |
| *ML Task* | Classification | Classification | Both contexts focus on the same ML task. (1) |
| *ML Model* | Convolutional LSTM Autoencoder | None | There exists no spatiotemporal model for target scenario. However, the ML model used for source task can be adapted once the AM model has been adapted. (1) |
| *ML Input* | Used: Graphic data of process melt pools in each layer of the printed specimen | Available: Graphic data of the process melt pools in each track of the printed sample | From ML perspective, there is a similarity in the process inputs (e.g., melt pool images) available to detect anomalies. (1) |
| *ML Preprocessing* | Applied: Miscellaneous preprocessing techniques, including cropping, noise reduction, scaling, and rotations | Applicable: all graphic transformations | The ML preprocessing used when developing source model can be applied to target data. (1) |
| *ML Output* | Binary (presence or absence of anomaly) | Binary (presence or absence of anomaly) | Output similarity leads to label space similarity for both tasks. (1) |

The detailed experimental settings used to produce LPBF data can be found in the description of dataset. The experiment was carried out on Additive Manufacturing Metrology Testbed (AMMT) at NIST, an open LPBF system. The experiment results in a geometry (5mm x 9mm x 5mm) on wrought nickel alloy 625 (IN625) plate. The process parameters (power and speed) for pre-contour and infill hatching were (100 W, 900 mm/s) and (195 W, 800 mm/s) respectively. A total of 250 layers each 20μm thick were printed with 90° rotation between layers. An optically aligned coaxial camera was used to monitor the melt pool. The camera specifications are detailed in Table 2.

In the case of DED process, A thin wall sample from Ti4Al6V powder was printed using an OPTOMEC LENS™ 750 system equipped with a 1 kW Nd:YAG laser (IPG) at MSU. The set dimensions (L x H) for the thin wall sample were 50.8 mm x 30.48 mm. The actual measured dimensions varied slightly. The substrate made of Ti4Al6V had dimensions of 153 x 153 x 3.3 mm$^3$. The processing parameters used for the fabrication were 290 W laser power, 12.7 mm/s scan rate, and the powder was fed at 0.32 g/s. A total of 60 tracks were printed with an upward increment of 0.508 mm. The overall process lasted seven minutes and thirty-seven seconds. The melt pool was recorded with pyrometer camera. The technical specifications are detailed in the Table 2.

Based on the framework, the pre-transfer analysis is carried out first. Table 2 comprehensively compares the identified knowledge components of the source, which is LPBF in this case study. The target is DED due to the lack of existing ML-based knowledge. Based on the similarity index (S=0.73), maturity factor (M=1) and availability factor (A=1), a pre-transfer score of 0.73 is obtained highlighting significant potential to knowledge transfer. In the ideal scenario, a comprehensive analysis with the existing literature may be conducted to identify all potential sources of knowledge.

The process to identify the TL model is carried out following steps of Figure 2. Similarities are observed in both AM domain as well as AM task. In AM domain, ML input (e.g., images) has the same representation allowing similar ML models to be developed in both cases. However, process (LPBF vs DED), material (Inconel vs Ti4Al6V) and system (Customized vs OPTOMEC LENS™ 750) differ leading to differences in marginal distribution. In AM task, an overlap is observed for both label space (e.g., anomalies) and conditional distribution (process activity and anomaly concern).

The pre-transfer step highlights significant transfer potential (e.g., pre-transfer score = 0.73) from source whereas domain and task comparisons indicate domain dissimilarity and task similarity (e.g., transductive learning). The applicable method to handle this scenario from TL literature is referred to as model or parameter-based transfer learning. This implies that the source and target scenarios share some common knowledge at the ML model level and the goal then becomes to exploit those common elements of source knowledge leading to the development of target data-driven model.

The datasets from source and target represent melt pool images captured in a time sequence. Figure 3 shows the source (A) and target (B) melt pools in a sequence taken randomly from the datasets. The source images represent the original LPBF melt pools in a 120 by 120 pixel window with greyscale representation. The target images represent processed DED melt pools in a much smaller window than the original size of 752 by 480. The processing represents their conversion from RBG to greyscale and subsequent cropping. This is done to bring both raw datasets in similar representation where the ML model can learn solely from the intensity gradients of pixels. Details on the source and target data can be found in [14] and [15] respectively. Later, both datasets are pre-processed to meet the input representation of the source model as closely as possible.

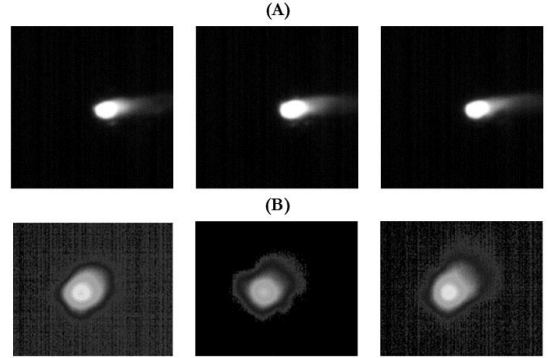

**FIGURE 3:** NORMAL MELT POOL EXAMPLES FROM LPBF (A) AND DED (B) IN TIME ORDER FROM LEFT TO RIGHT

As a pre-requisite to model or parameter-based transfer, the datasets from source and target context are represented in the spatiotemporal structure proposed by Ko et al. [4]. Specifically, the build process can be decomposed into inner-layer and layer-wise transitions to jointly represent the resulting build. Let $x_i$ represents the states of neighboring layers in the build, then the build can be represented as the concatenation with a series of $L$ states representing neighboring layer as

$$x = \prod_{i=1}^{L} x_i, 1 \leq L < \infty \tag{7}$$

The inner layer transitions are defined as those by control and time advance. As a result, the layer state $x_i$ can be decomposed into the controls that contribute to the completion of that layer via inner states $x_{i,j}$, represented by $M$ in total. This is represented as

$$x_i = \prod_{j=1}^{M} x_{i,j}, 1 \leq M < \infty \tag{8}$$

Each control step can be further decomposed in terms of the transitions $x_{i,j,k}$ resulting from time advance. For a given control step $x_{i,j}$, this can be represented as follows for a total of $N$ time advances involved:

$$x_{i,j} = \prod_{K=1}^{N} x_{i,j,k}, 1 \leq N < \infty \tag{9}$$

Finally, to include both layer-wise and inner-layer transitions, the build dataset is represented as Equation (10):

$$\prod_{i=1}^{L} \prod_{j=1}^{M} \prod_{k=1}^{N} x_{i,j,k}, 1 \leq L, M, N < \infty \tag{10}$$

The LPBF and DED datasets are structured following the framework described above. Specifically, the spatial representations of individual melt pools are concatenated in a temporal order leading to spatiotemporal concatenations for ML model inputs. We use a time window of 4 (e.g., four melt pools) and apply a sliding window to generate these concatenations from both source and target datasets. In the case of LPBF, only the frame belonging to one layer (210[th]) are used for training while the frames belonging to another layer (150[th]) are used for testing. Each source layer has large number (e.g., ~20,000 for 210[th] and ~16,000 for 150[th]) of melt pool images owing to the high frame capture rate of the camera. We used a subset of 5,000 frames from each layer for training and testing purposes. Since the frame capture rate in DED is much lower than LPBF, significantly less melt pool images are captured. As a result, almost the entire dataset representing the spatiotemporal depositions in the thin-wall sample is used. After removing the frames corresponding to laser off mode, we are left with approximately 1578 melt pool frames, representing a 1:3 of DED to LPBF images. Out of these frames, 1047 normal frames are used for training representing almost a ratio of 1 to 5 for DED to LPBF images used for training. The total anomaly images available for test are 148. This scenario represents a practical situation for applying TL to exchange knowledge from data rich source to data scarce target.

The anomaly detection model requires a criterion for filtering the anomalous concatenations from the normal ones to support the training process. The process of transferring knowledge is expected to be independent of the condition of having the same anomaly criteria between source and target. This allows domain and application specific criteria to be selected. In the case of LPBF, the anomalies are defined as the presence of noise, plume, and spatter. Figure 4 (A) shows some examples of LPBF melt pool anomalies. In the case of DED, the anomalies are defined based on irregular melt pool shapes such as the ones shown in part B of the Figure 4. The exact criteria to process and filter the anomalous concatenations in both cases are refined through trial and error involving visual validation and manual selection.

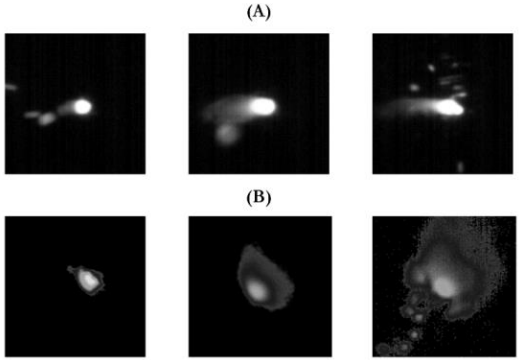

**FIGURE 4:** ANOMALOUS MELT POOL EXAMPLES FROM LPBF (A) AND DED (B) PICKED RANDOMLY

Once the data is processed and structured, we first reproduce the implementation of source ML model as per the details in [4]. The ML model being reimplemented and later considered for TL is Convolutional Long Short-Term Memory (LSTM) Autoencoder normally deployed for video anomaly detection tasks. Table 3 presents architectural details of the model. The unique aspect in this kind of anomaly detection model is the bottleneck layer that preserves both spatial and temporal dependencies in a lower dimension. This is accomplished by Convolutional LSTM layer which learns both spatial and temporal features [16]. Equations (11) to Equation (16) usually represent a convolutional LSTM layer.

$$i_t = \sigma(W_{Xi} * X_t + W_{Hi} * H_{t-1} + W_{ci} \odot C_{t-1} + b_{Hi}) \quad (11)$$
$$f_t = \sigma(W_{Xf} * X_t + W_{Hf} * H_{t-1} + W_{cf} \odot C_{t-1} + b_f) \quad (12)$$
$$g_t = \tanh(W_{Xg} * X_t + W_{Hg} * H_{t-1} + b_{h-g}) \quad (13)$$
$$C_t = f_t \odot C_{t-1} + i_t \odot g_t \quad (14)$$
$$o_t = \sigma(W_{Xo} * X_t + W_{Ho} * H_{t-1} + W_{Co} \odot C_t + b_{f-o}) \quad (15)$$
$$\mathcal{H}_t = o_t \odot \tanh(C_t) \quad (16)$$

As per the original article on Convolutional LSTM or ConvLSTM, the network preserves the convolutional structure in both input to state and state to state transitions [16]. The future states are predicted based on the previous states and the input. The key equations in this regard are presented above with $X$ representing the data, $W$ the kernels, $b$ the biases, $C$ the cell outputs, $H$ the hidden states, and $i$, $f$ and $o$ as the input, forget, and the output gates respectively. For the operations, $*$, $\odot$, $\sigma$, and tanh represent convolutions, Hadamard product, sigmoid, and hyperbolic tangent functions.

**TABLE 3:** SOURCE MODEL ARCHITECTURE. ALL LEARNABLE LAYERS ARE TIME DISTRIBUTED AND PERFORM FRAME-WISE 2D OPERATIONS. THE ENCODER (BLUE), DECODER (ORANGE) AND SPATIOTEMPORAL BOTTLENECK (GREEN) ARE HIGHLIGHTED. "BN" REPRESENTS BATCH NORMALIZATION AND "T" INDICATES TRANSPOSE OPERATION

| Layers | Channel (in/out) | Kernel Size | Stride | Activation |
|---|---|---|---|---|
| Conv_2D | 1/128 | 5 by 5 | 2 | Relu |
| BN | 128/128 | | | |
| Conv_2D | 128/64 | 5 by 5 | 2 | Relu |
| BN | 64/64 | | | |
| ConvLSTM | 64/64 | 3 by 3 | 1 | Relu |
| BN | 64/64 | | | |
| ConvLSTM | 64/32 | 3 by 3 | 1 | Relu |
| ConvLSTM | 32/64 | 3 by 3 | 1 | Relu |
| BN | 64/64 | | | |
| Conv_2D_T | 64/64 | 5 by 5 | 2 | Relu |
| BN | 64/64 | | | |
| Conv_2D_T | 64/128 | 5 by 5 | 2 | Relu |
| BN | 128/128 | | | |
| Conv_2D_T | 128/1 | 2 by 2 | 1 | Sigmoid |

We reproduce the model with Keras and associated libraries. While the chosen source melt pool images and subsequent preprocessing steps vary in the reimplementation, same hyperparameters (e.g., epochs, loss function, optimizer) are used to train the model as input distribution and domain remains the same. This also leads to a difference in the performance of reimplemented source model.

TL is conducted after the source model carrying the LPBF melt pool anomaly detection knowledge is obtained. Transferability analysis justifies the knowledge transfer from the source to target tasks with high similarities such as the task and data. It also highlights the key geometric and temporal differences to guide TL. Geometrically, the sizes of the melt pools from the two datasets are different dictated by different process scale and measurement devices: temporally, different frame rates and solidification speeds yields different time-series patterns. Thus, both the CNN and ConvLSTM layers must be re-trained to adapt to the target task. In such as deep and complex source model, it has the risk of 'catastrophic forgetting' if all layers are re-trained simultaneously [17]. Overwhelmed by the re-training process, the model might fail to adapt to the target task and even lose the source knowledge. Therefore, three re-training strategies are applied to investigate and avoid negative transfer. As a result, three levels of source knowledge (e.g., data representation, data structuring/processing, and model parameters) eventually get transferred to target's context. Figure 5 highlights different levels of transferred source knowledge.

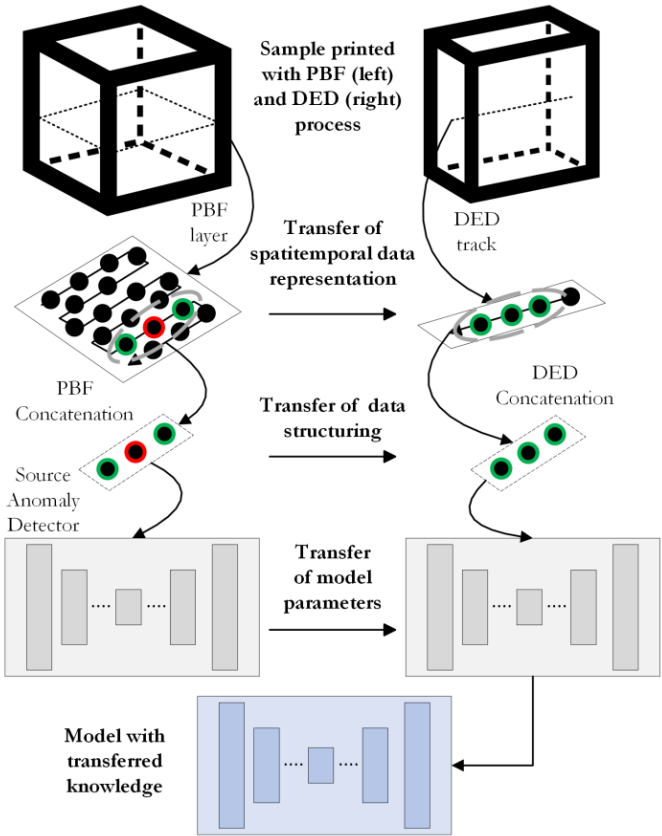

**FIGURE 5:** REGULARITY SCORES FOR 200 DED CONCATENATIONS WITH DETECTED ANOMALIES

## 4. RESULTS AND DISCUSSION

To test the performance on video anomaly detection task, we use a regularity score to detect concatenations with irregular or anomalous frames. Then, a threshold is chosen to filter the normal concatenations having higher regularity score from those having lower scores. The accuracy of the autoencoder is defined as the percentage of detected anomalies from all anomalies in the test data. The regularity score first computes pixel-wise intensity differences between original and reconstructed frames at each timestamp of a given concatenation. This is done using L2 Norm. The reconstruction error for a given frame is computed by summing all pixel-wise differences. Later, the frame-wise differences are summed across a concatenation to get its reconstruction cost. The reconstruction costs of all concatenations in the train dataset are normalized between 0 and 1, defined as the abnormality score. Finally, the regularity scores are obtained by subtracting abnormality scores from 1. The step to calculate regularity scores are shown in Equations (17) to (21).

$$d(x,y) = \|O(x,y) - f_{ae}(O(x,y))\| \quad (17)$$
$$d_t = \sum_{(x,y)} d(x,y) \quad (18)$$
$$r_t = \sum_t^{t+4} d_t \quad (19)$$
$$sa_t = (r_t - r_{min})/r_{max} \quad (20)$$
$$sr_t = 1 - sa_t \quad (21)$$

where $d(x,y)$ matrix represents pixel-wise differences within an image. $d_t$ represents sum of errors over the entire image, $r_t$ represents sum or errors over the entire concatenation. $sa_t$ is the normalized abnormality score and $sr_t$ is the regularity score.

The reimplemented model on source dataset from NIST obtained an accuracy of 90% when tested on unseen data from a different layer. To have a fair comparison with the source performance, we used reconstruction error instead of the final regularity scores and chose the maximum value on the train set as the threshold (e.g., source uses 99[th] percentile of anomaly scores on train set). The difference in performance from reported results (e.g., 98%) can be explained on several factors such as preprocessing, anomaly metrics, and dataset selection. First, we only normalized the raw dataset as opposed to augmentations (e.g., rotation and scaling) that were done in the source context to improve the performance and avoid overfitting. Secondly, the choice of dataset varies in two aspects as the exact images and anomaly criteria chosen for re-implementation are different. The labeling of source dataset was done manually to filter different types of abnormalities. This requires significant effort. While preparing the data to train with normal concatenations, we only filtered images with spatter, plume or obvious noises leaving the irregular melt pool shapes in the set. This reimplemented model was considered good enough to test transfer learning scenarios from source to target on anomaly detection task.

The source architecture is trained on target DED data from scratch to see how well it performs on the test data before the transfer learning is carried out. A total of 1,047 concatenations are used for train set while total available anomalous concatenations constituting the test set for the target are 148. Since regularity score differs from reconstruction error, minimum regularity score on train set can be selected as opposed to maximum reconstruction error. This ensures that threshold accounts for all regular images. We choose 3[rd] minimum regularity score on train set to avoid any noisy outlier. Without TL, the trained and optimized source architecture performs 84% accurately on the target test set when trained from scratch. Figure 6 shows the regularity scores for a subset of target data showing three anomaly images with their low regularity scores.

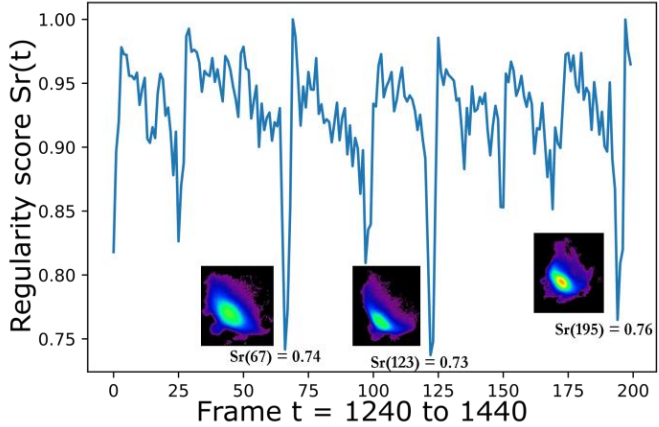

**FIGURE 6:** REGULARITY SCORES FOR 200 DED CONCATENATIONS WITH DETECTED ANOMALIES

A total of three TL strategies are used to fine tune reimplemented source model of LPBF data on target DED data. In the first strategy, all layers are re-trained simultaneously for 200 epochs. The second strategy first freezes the CNN layers while re-training the ConvLSTM layers for 100 epochs, thereafter, freezes the ConvLSTM layers while re-training the CNN layers for the next 100 epochs. The third strategy reverses the sequence of the second strategy by first re-training the CNN layers then re-training the ConvLSTM layers. Figure 7 shows significant reduction in loss value during the training process which is not achievable when the model is trained solely on the target data. The three TL strategies appear to converge at the same loss value.

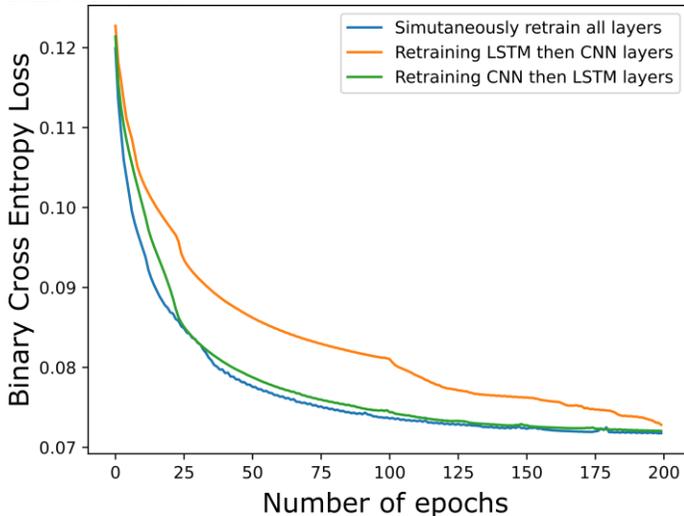

**FIGURE 7:** ERROR CURVES FROM TRANSFER LEARNING STRATEGIES. THE FINE-TUNING LEADS TO SIGNIFICANT REDUCTION OF ERROR FROM MINIMUM POSSIBLE WITHOUT TL

**TABLE 4:** ANOMALY DETECTION RESULTS ON TARGET DED DATA WITH AND WITHOUT TL. THE MAXIMUM ACCURACY IN EACH SETTING IS EMBOLDEN

| Scenario | No TL | Retrain All | Retrain CNN, Retrain ConvLSTM | Retrain ConvLSTM, Retrain CNN |
|---|---|---|---|---|
| Accuracy | 84 | 84 → **94** | 84→90, 90→**93** | 84→ 91, 91→ **94** |

Table 4 indicates the performance of each TL strategy against training from scratch. Without TL, the model achieves 84% performance on the DED test data. TL is deployed to overcome data limitation and lower performance. The reimplemented source model acts as the starting point. When all its layers are fine-tuned on target DED data, the model performance improves from 84% to 94%, a significant 10% improvement. In the later scenarios, retraining is split between pure spatial and spatiotemporal layers. Both settings lead to similar performance to the scenario of retraining all layers. The slight improvement in the case of retraining ConvLSTM first can be related to the ability of these layers to learn both spatial and temporal features simultaneously. It is interesting to note that the first step in each setting leads to most of the improvement in performance which is also evident from learning curves of Figure 7 where first 100 epochs represent majority of the error drop on train set.

We made two important observations during the process of TL. Since most of the anomalous DED concatenations consist of only one anomaly image, we note that excessive fine-tuning can lead to the regularization of anomalies. As the model learns normal features during the TL, the reconstruction cost decreases significantly. When the model is tested on anomalous concatenations, the net regularity score of each concatenation improves, giving the notion of regularity. This can be avoided by limiting the transfer (e.g., lower number of epochs) or calculating regularity scores solely based on the anomaly images of a concatenation. Also, the process images from AM are not as information rich as real-world images. As a result, there exist smaller differences in the regularity scores of anomalous and normal concatenations, which requires careful selection of the threshold.

The current validation of the method is limited to metal AM. Similarly, the knowledge components are less quantitative and lack objectiveness in their existing representations.

## 5. CONCLUSION

Inspired by the increasing frequency of research into data-driven solutions of AM challenges, we propose a knowledge transferability analysis framework. The three-step framework compares knowledge components of different AM solutions to determine the nature of transferability possible. As a pre-requite to the knowledge transferability analysis, key components of AM and ML knowledge are identified to represent data-driven AM knowledge. A pre-transfer analysis is used to highlight potential and applicable sources of knowledge. We link the nature of similarity to the type of transfer possible and validate post-transfer to update the existing AM knowledge. A case study is conducted between knowledge rich source and knowledge scarce target AM contexts. We successfully adapt source data representation, ML model architecture, and ML model parameters to target context. In the future, the pre-transfer process can be automated to allow efficient and comprehensive comparison with the existing literature. The identified knowledge components can be made robust to enable better transferability analysis.

## ACKNOWLEDGEMENTS

McGill Engineering Doctoral Award (MEDA) fellowship for Mutahar Safdar is acknowledged with gratitude. Mutahar Safdar also received financial support from National Research Council of Canada (Grant# NRC INT-015-1). McGill Graduate Excellence Award (Grant# 00157), Mitacs Accelerate Program (Grant# IT13369), and MEDA fellowship for Jiarui Xie are acknowledged with gratitude. The authors are grateful to Digital Research Alliance of Canada (RRG# 4294) for providing computational resources to support this research. We acknowledge the open availability of data from NIST and Mississippi State University that made this research possible.